\def\year{2021}\relax
\documentclass[letterpaper]{article} % DO NOT CHANGE THIS
\usepackage{aaai21}  % DO NOT CHANGE THIS
\usepackage{times}  % DO NOT CHANGE THIS
\usepackage{helvet} % DO NOT CHANGE THIS
\usepackage{courier}  % DO NOT CHANGE THIS
\usepackage[hyphens]{url}  % DO NOT CHANGE THIS
\usepackage{bm}
\usepackage{multirow}
\usepackage{threeparttable}
\usepackage{graphicx} % DO NOT CHANGE THIS
\usepackage{natbib}  % DO NOT CHANGE THIS AND DO NOT ADD ANY OPTIONS TO IT
\usepackage{caption} % DO NOT CHANGE THIS AND DO NOT ADD ANY OPTIONS TO IT
\usepackage{color}

\title{
Leveraging Domain Agnostic and Specific Knowledge for Acronym Disambiguation
}
\author{
    Qiwei Zhong, Guanxiong Zeng, Danqing Zhu, Yang Zhang, Wangli Lin, Ben Chen, Jiayu Tang\\
}
\affiliations{
    Alibaba Group, Hangzhou, China\\
    \{yunwei.zqw, moshi.zgx, danqing.zdq, zy142206, wangli.lwl, chenben.cb, jiayu.tangjy\}@alibaba-inc.com

}
\begin{document}

\maketitle
\begin{abstract}
An obstacle to scientific document understanding is the extensive use of acronyms which are shortened forms of long technical phrases. Acronym disambiguation aims to find the correct meaning of an ambiguous acronym in a given text. Recent efforts attempted to incorporate word embeddings and deep learning architectures, and achieved significant effects in this task.
In general domains, kinds of fine-grained pretrained language models
have sprung up, thanks to the large-scale corpora which can usually be obtained through crowd-sourcing.
However, these models based on domain agnostic knowledge might achieve insufficient performance when directly applied to the scientific domain. Moreover, obtaining large-scale high-quality annotated data and representing high-level semantics in the scientific domain is challenging and expensive.
In this paper, we consider both the domain agnostic and specific knowledge, and propose a \underline{H}ierarchical \underline{D}ual-path \underline{BERT} method coined \textbf{hdBERT} to capture the general fine-grained and high-level specific representations for acronym disambiguation. First, the context-based pretrained models, RoBERTa and SciBERT, are elaborately involved in encoding these two kinds of knowledge respectively. Second, multiple layer perceptron is devised to integrate the dual-path representations simultaneously and outputs the prediction. With a widely adopted SciAD dataset contained 62,441 sentences, we investigate the effectiveness of hdBERT. The experimental results exhibit that the proposed approach outperforms state-of-the-art methods among various evaluation metrics. Specifically, its macro F1 achieves 93.73\%.

\end{abstract}

\section{Introduction}
\label{sec:introduction}

\begin{table*}
    \centering
    \resizebox{\linewidth}{!}{
    \begin{tabular}{ll}
    \hline
         \bm{\mathrm{Input - Sentence}}: & They use \bm{\mathrm{CNN}} in the proposed model.\\
\bm{\mathrm{Input - Dictionary}}: & CNN: 1. Convolutional Neural Network, 2. Cable News Network, 3. Condensed Nearest Neighbor\\
\hline
\bm{\mathrm{Output}}: & Convolutional Neural Network\\
\hline
    \end{tabular}
    }
    \caption{A toy sample of acronym disambiguation.}
    \label{tab:example}
\end{table*}

In recent years, it has witnessed the vigorous development of deep learning.
Among the most successful scenarios, natural language processing (NLP) is advancing steadily.
However, natural language is frequently ambiguous, so many words and phrases can be interpreted in many ways depending on the context in which they appear~\cite{navigli2009word}.
Specifically, an obstacle to scientific document understanding (SDU) is the widespread use of acronyms, which are shortened forms of long technical phrases~\cite{veyseh-et-al-2020-what,Beltagy2019SciBERT}.
In order to understand the document correctly, the SDU system should be able to identify acronyms and their correct meanings.
The goal of acronym disambiguation (AD) is to determine the correct long form of an ambiguous acronym in a given text~\cite{veyseh2020acronym}.
%~\footnote{We won second place in the acronym disambiguation competition. \url{https://sites.google.com/view/sdu-aaai21/shared-task}}.
It is usually formulated as a sequence classification problem in general~\cite{veyseh-et-al-2020-what}.
For instance, a toy sample of this task is shown in Table~\ref{tab:example}.
In this example, 
the ``\textit{CNN}'' might be an acronym for ``\textit{Convolutional} \textit{Neural} \textit{Network}'', ``\textit{Cable} \textit{News} \textit{Network}'' or ``\textit{Condensed} \textit{Nearest} \textit{Neighbor}''.
Given a sentence ``\textit{They} \textit{use} \textit{\textbf{CNN}} \textit{in} \textit{the} \textit{proposed} \textit{model}.'' and a dictionary with possible expansions (i.e., long forms) of the acronym ``\textit{CNN}'', the expected prediction for its correct meaning is ``\textit{Convolutional} \textit{Neural} \textit{Network}''.
Recent efforts attempted to incorporate hand crafted features~\cite{li2018guess}, word embeddings~\cite{charbonnier2018using,ciosici2019unsupervised}, graph structures~\cite{prokofyev2013ontology,veyseh-et-al-2020-what}, and deep learning architectures~\cite{jin2019deep,blevins2020moving} and achieved significant effects in this task.

In this paper, we pay more attention to the scenario of scientific acronym disambiguation.
Some observations are still worthy of further investigation.
Generally, large-scale training data for natural language processing tasks in general domains is often possible to obtain through crowd-sourcing, emerging a variety of domain-independent fine-grained pretrained models.
However, these models based on domain agnostic knowledge might achieve insufficient performance when applied to the specific domain~\cite{Beltagy2019SciBERT}.
Furthermore, obtaining large-scale annotated data in the scientific domain is challenging and expensive~\cite{Beltagy2019SciBERT}, which leads to the shortage of high-level semantic expression to some extent.

To remedy these challenges, we fully consider both the domain agnostic and specific knowledge, and propose a \underline{H}ierarchical \underline{D}ual-path \underline{BERT} method coined \textbf{hdBERT} to fusion the general fine-grained and high-level specific representations for acronym disambiguation.
The overall architecture is illustrated in Figure~\ref{fig:structure}.
We pinpoint that hdBERT is a BERT-based supervised method adopting the now ubiquitous transformer architecture~\cite{vaswani2017attention}.
First, RoBERTa~\cite{liu2019roberta} and SciBERT~\cite{Beltagy2019SciBERT} modules are elaborately involved to distill representations from inputs consist of sentence and candidate long forms.
Specifically, we utilize RoBERTa, a robustly optimized method trained on general domain corpora via byte-level Byte-Pair-Encoding~\cite{sennrich2016neural}, to capture domain agnostic and fine-grained semantic information.
Moreover, SciBERT which is also a pretrained language model based on BERT~\cite{devlin2018bert} is exploited to model the high-level scientific domain representation.
Since it leverages unsupervised pretraining on a large multi-domain corpus of scientific publications using WordPiece~\cite{wu2016google} tokenization strategy.
Second, we integrate these dual-path representations from RoBERTa and SciBERT simultaneously via multiple layer perceptron and output the prediction. %Finally, the model output the prediction result.
The main contributions of this work are summarized as follows:
\begin{itemize}
	\item We are the very first attempt to resolve the acronym disambiguation problem simultaneously leveraging domain agnostic and specific knowledge.
	\item We propose a novel hierarchical dual-path BERT method coined hdBERT to capture both general fine-grained and high-level specific representations. It is mainly implemented based on the well-known transformer architecture, which can train the overall model more effectively.
	\item Experiments on real-world datasets demonstrate the effectiveness of the proposed approach. It achieves competitive performance and outperforms state-of-the-art methods.
\end{itemize}

\section{Related Work}
\label{sec:relatedwork}
In this section, we review the related researches on word sense disambiguation especially acronym disambiguation as well as BERT and its two representative variants.

\subsection{Word Sense Disambiguation}
Word sense disambiguation (WSD) is an open problem concerned with identifying which sense of a word is used in a text~\cite{navigli2009word}.
It is a core and difficulty in natural language processing tasks, which affects the performance of almost all downstream tasks.
The methods to solve word sense disambiguation are usually divided into two categories: knowledge-based and supervised~\cite{wang2020word,barba2020mulan}.

Knowledge-based methods usually rely on amounts of statistical information and can be easily extended to other low-resource languages~\cite{agirre2014random,scarlini2020sensembert}.
For example, SensEmBERT~\citep{scarlini2020sensembert}, a knowledge- and BERT-based method that combines the expressive power of language modeling with the vast amount of knowledge contained in the semantic network, produces high-quality latent semantic representations of the meanings of the word in different languages.
And it can achieve competitive results attained by most of the supervised neural approaches on the WSD tasks.
On the other hand, supervised methods require lots of labeled data to learn word representations~\cite{bevilacqua2020breaking,wang2020word}.
Of course, this defect can be alleviated through semi-supervised methods~\cite{barba2020mulan} by jointly leveraging contextualized word embedding and the multilingual information to project some sense labels.

Furthermore, acronym disambiguation is more challenging since we need to identify the acronym first and then to understand the text to determine the correct meaning of acronyms.
Recently, an effective solution is to extract acronym definitions from unstructured texts by computing the Levenshtein string edit distance between any pair of long forms~\cite{ciosici2019unsupervised}, which is an entirely unsupervised acronym disambiguation method. And researches also attempt to incorporate hand crafted features~\cite{li2018guess}, word embeddings~\cite{charbonnier2018using,ciosici2019unsupervised}, graph structures~\cite{prokofyev2013ontology,veyseh-et-al-2020-what}, and deep learning architectures~\cite{jin2019deep,blevins2020moving}, and have achieved significant effects in this task.
Specifically, a supervised method named GAD~\cite{veyseh-et-al-2020-what}, which utilizes the syntactic structure of sentences to extend ambiguous acronyms in sentences by combining Bidirectional
Long Short-Term Memory~(BiLSTM) with Graph Convolutional Networks~(GCN), provides a strong baseline on acronym disambiguation tasks in the scientific domain.
\begin{figure*}
    \centering
    \includegraphics[width=\linewidth]{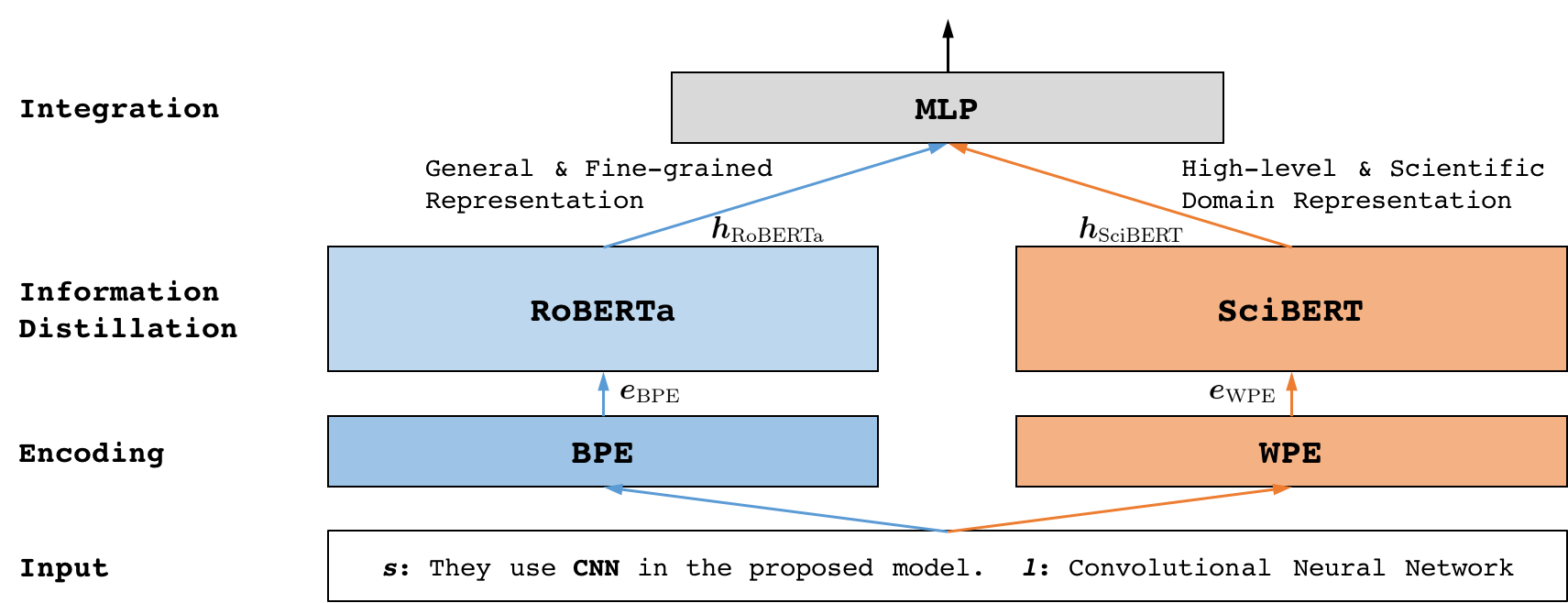}
    \caption{Illustration of the proposed hdBERT model.}
    \label{fig:structure}
\end{figure*}
%The computational recognition of the meaning of a word in the context is called word sense disambiguation~(WSD)~\cite{navigli2009word}.
%WSD is a core and difficult point in natural language processing tasks, which affects the performance of almost all tasks.
%But it's relatively difficult for the machine to process unstructured text information and convert it into a data structure to determine the potential meaning.
\subsection{BERT-based Methods}
Bidirectional Encoder Representations from Transformers (BERT) \cite{devlin2018bert} is a self-supervised learning method that trains based on a large number of corpora to express better features for word embedding. And its network architecture utilizes the multi-layer transformer structure~\cite{vaswani2017attention}.
The feature representation of BERT could be directly adopted as word embedding features for downstream tasks.
Besides, BERT provides a model for transfer learning of other tasks. It can be fine-tuned or fixed according to tasks and then treated as a feature extractor.
BERT was significantly undertrained, and there have been many fine-grained improvements or specific domain variants of it~\cite{Beltagy2019SciBERT,liu2019roberta,scarlini2020sensembert,lee2020biobert}.
\subsubsection{RoBERTa.}
RoBERTa~\cite{liu2019roberta} is mainly trained on general domain corpora via byte-level Byte-Pair-Encoding \cite{sennrich2016neural} based on the structure of BERT and can supply more fine-grained representation.
This encoding scheme can process amounts of words that are common in natural language corpora and is more conducive to the translation of acronyms.

\subsubsection{SciBERT.}

SciBERT~\cite{Beltagy2019SciBERT} is a specific pretrained language model for scientific domain texts. This model follows the same architecture as BERT to solve the lack of high-quality, large-scale labeled scientific data. It significantly outperforms previous BERT-based methods and achieves new state-of-the-art results on some scientific NLP tasks.

\section{Methodology}
\label{sec:methodology}
In this section, we first introduce the problem statement of acronym disambiguation and then describe the overall architecture and details of our proposed hdBERT model.
\subsection{Problem Statement}
Acronym disambiguation is formulated as a sequence classification problem in general~\cite{veyseh-et-al-2020-what}.
Formally, given an input sentence $\bm{s}=w_1, w_2, ..., w_n$ and the position of the acronym, i.e., $p$, the goal is to disambiguate the acronym $w_p$, that is, predicting the true long form $\bm{l}$ from all candidate long forms of $w_p$.
Specifically, in this paper, we simplify it into a binary classification problem. That is, given an input sample consists of the sentence $\bm{s}$ with acronym $w_p$ and the candidate long form $\bm{l}$, i.e., $\bm{x}=(\bm{s}; \bm{l})$, our purpose is to predict the probability of $\bm{l}$ being the right long form of $w_p$. 
We assign a label $y \in \{0,1\}$ on each sample in training dataset to indicate whether $\bm{l}$ is a true long form of $w_p$ in sentence $\bm{s}$ or not.
In the testing phase, the long form with the highest prediction probability among the candidate long form set of a sentence would be chosen as its final result.

\subsection{Overview}
Figure~\ref{fig:structure} exhibits the schematic illustration of the proposed hdBERT model.
As mentioned previously, we design a hierarchical integration model comprising three major components, each plays a different role in final prediction.
The first two context-based components, i.e., RoBERTa~\cite{liu2019roberta} and SciBERT~\cite{Beltagy2019SciBERT} modules, distill representations of the sentence and the candidate long forms.
Specifically, as a robustly optimized method trained on vast amounts of general domain corpora, we use RoBERTa to capture the general and fine-grained semantic information via byte-level Byte-Pair-Encoding~\cite{sennrich2016neural}.
Moreover, SciBERT, which leverages unsupervised pretraining on a large scientific corpus by WordPiece~\cite{wu2016google} tokenization strategy, is exploited to represent the high-level scientific domain information.
Finally, a multiple layer perceptron network is devised to fusion these two kinds of representations. In the following, we present detail of each major component.

\subsection{Information Distillation}
\subsubsection{General and Fine-grained Information.}
We involve RoBERTa to capture domain agnostic and fine-grained information of the sentence and its candidate long form.
RoBERTa uses the now ubiquitous transformer architecture~\cite{vaswani2017attention} via byte-level Byte-Pair-Encoding (BPE), which is a hybrid between character- and word-level representations that allow handling large vocabularies common in natural language corpora. Instead of full words, BPE relies on subwords units, which are extracted by performing statistical analysis of the training corpus.
The size of the original vocabulary released with RoBERTa is about 50K, which is 20K more than BERT's.

We define the encoding of a sample $\bm{x}=(\bm{s},\bm{l})$ after the BPE strategy as $\bm{e}_{\mathrm{BPE}}$ and the output representation throughout the RoBERTa model as $\bm{h}_{\mathrm{RoBERTa}}$.
\begin{equation}
    \bm{e}_{\mathrm{BPE}} = \bm{\mathrm{BPE}}(\bm{x})
\end{equation}
\begin{equation}
    \bm{h}_{\mathrm{RoBERTa}} = \bm{\mathrm{RoBERTa}}(\bm{e}_{\mathrm{BPE}})
\end{equation}

\subsubsection{High-level Scientific Domain Information.}
To handle the high-level scientific domain information, SciBERT is chosen elaborately. SciBERT follows the same architecture as BERT but is instead pretrained on the scientific texts.
% illustrating a substantial difference in frequently used words between scientific and general domain texts.
It constructed a new WordPiece vocabulary on scientific corpus using the SentencePiece
%\footnote{\url{https://github.com/google/sentencepiece}} 
library and trained on a random sample of 1.14M papers from Semantic Scholar~\cite{ammar2018construction}. Its corpus consists of 18\% papers from the computer science domain and 82\% from the broad biomedical domain.
The size of the original vocabulary released with SciBERT is about 30K, which is 20K less than RoBERTa. The resulting token overlap between SciBERT and BERT is 42\%,
which illustrates the significant difference in common terms between scientific and general domain texts.

We define the encoding of a sample $\bm{x}=(\bm{s},\bm{l})$ after SciBERT's encoding strategy (noted as WPE) as $\bm{e}_{\mathrm{WPE}}$ and the output representation throughout the SciBERT model as $\bm{h}_{\mathrm{SciBERT}}$.
\begin{equation}
    \bm{e}_{\mathrm{WPE}} = \bm{\mathrm{WPE}}(\bm{x})
\end{equation}
\begin{equation}
    \bm{h}_{\mathrm{SciBERT}} = \bm{\mathrm{SciBERT}}(\bm{e}_{\mathrm{WPE}})
\end{equation}

\subsection{Integration}
After modeling the two complex representations above, the obtained concatenation $\bm{h}$ is fed into multiple layer perceptron network and followed by a regression layer with $\mathrm{sigmoid}$ unit, as follows:
\begin{equation}
    \bm{h}=[\bm{h}_{\mathrm{RoBERTa}};\bm{h}_{\mathrm{SciBERT}}]
\end{equation}
\begin{equation}
p = \mathrm{sigmoid}(\bm{W}^\mathrm{T} \bm{\mathrm{MLP}}(\bm{h}) + b)
\end{equation}
where $\bm{W}$ is the weight vector, $b$ is the bias, and $\bm{\mathrm{MLP}}(\cdot)$ represents the operation of multiple layer perceptron shown in Figure~\ref{fig:structure}. Here $p$ is the predicted probability.

Finally, our model is trained with cross entropy loss with regularization. The loss function is defined as
\begin{equation}
\label{eq:loss}
\mathcal{\bm{L}}(\theta) = -\sum_{\mathcal{D}}{\left(y\log(p)+(1-y)\log(1-p)\right)} + \lambda{\|\bm{\theta}\|}_2^2
\end{equation}
where $y$ is the ground truth, $\bm{\theta}$ is the parameter set of the proposed model, $\lambda$ is the regularizer parameter, and $\mathcal{D}$ is the training dataset.
\section{Experiments}
\label{sec:experiments}
\begin{table}
    \centering
    \begin{tabular}{l|c}
    \hline
        \bm{\mathrm{Statistical Information}} & \bm{\mathrm{SciAD}} \\
         \hline
        number of acronyms & 732 \\
        average number of long form per acronym & 3.1 \\
        overlap between sentence and long forms & 0.32 \\
        average sentence length & 30.7 \\
         \hline
         number of training & 50,034\\
         number of development & 6,189\\
         number of test & 6,218\\
         \hline
    \end{tabular}
    \caption{The statistical information of original SciAD dataset. Note that the third row shows the ratio of sentences that have at least one word in common with the long forms of the acronyms appearing in the sentence.}
    \label{tab:statinfo}
\end{table}

In this section, we first illustrate the datasets, evaluation metrics, and implementation details, then demonstrate the experimental results and further studies.

\subsection{Datasets}
\begin{figure}
    \centering
    \includegraphics[width=\linewidth]{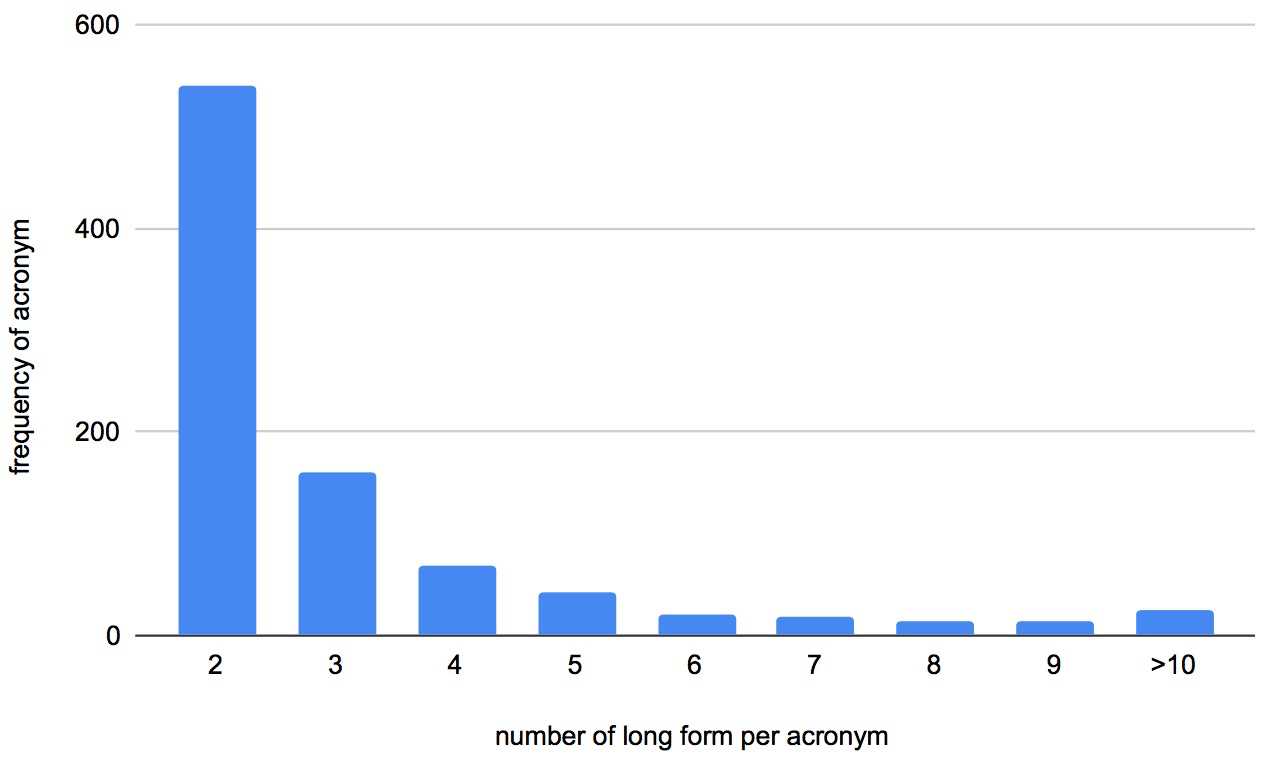}
    \caption{Distribution of acronyms based on number of long form per acronym.}
    \label{fig:disofacr}
\end{figure}

\begin{figure}
    \centering
    \includegraphics[width=\linewidth]{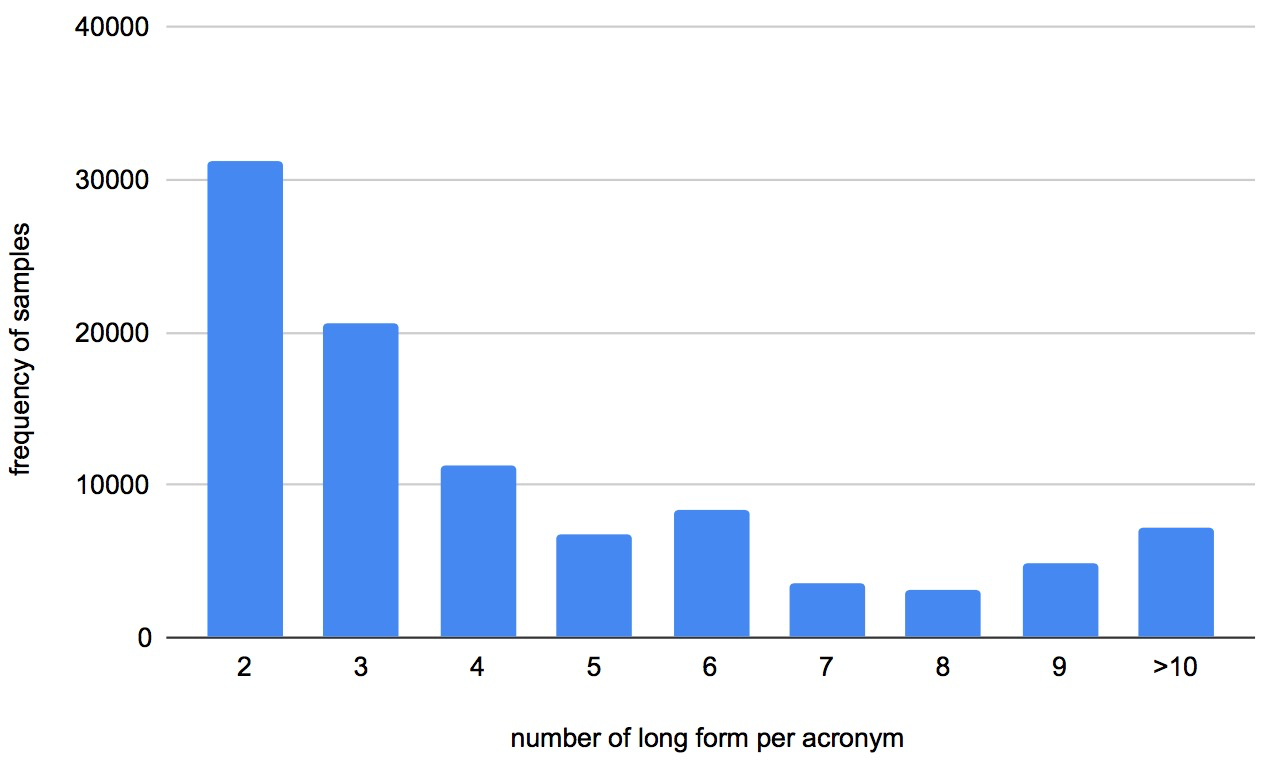}
    \caption{Distribution of samples based on number of long form per acronym.}
    \label{fig:disofsample}
\end{figure}

The SciAD~\footnote{We won second place in the acronym disambiguation competition. \url{https://sites.google.com/view/sdu-aaai21/shared-task}} dataset created from 6,786 English scientific papers
aims to find the correct meaning of an ambiguous acronym in a given sentence~\cite{veyseh-et-al-2020-what}. It contains 62,441 sentences and a dictionary of 732 ambiguous acronyms.
More statistical information is shown in Table~\ref{tab:statinfo}.
Besides, a toy sample of the SciAD dataset is shown in Table~\ref{tab:example}. The input is a sentence with an ambiguous acronym and a dictionary with possible expansions (i.e., long forms) of the acronym. In this example, the ambiguous acronym ``\textit{CNN}'' in the input sentence is shown in boldface and the expected prediction for its correct meaning is ``\textit{Convolutional} \textit{Neural} \textit{Network}''.
In addition, Figures~\ref{fig:disofacr} and~\ref{fig:disofsample} demonstrate more statistics of SciAD dataset~\cite{veyseh-et-al-2020-what}. More specifically, Figure~\ref{fig:disofacr} shows the distribution of the number of acronyms based on the number of long forms per acronym, and the distribution of the number of samples based on the number of long form per acronym is shown in Figure~\ref{fig:disofsample}.

\begin{table*}
    \centering
    \resizebox{\linewidth}{!}{
    \begin{threeparttable}
    \begin{tabular}{l|ccc}
    \hline
         \textbf{Parameter} & \textbf{BERT} & \textbf{RoBERTa} & \textbf{SciBERT} \\
         \hline
         pretrained model & bert-large-ucased\tnote{a} & roberta-large\tnote{b} & allenai/scibert\_scivocab\_uncased\tnote{c} \\
         architecture & sequence classification & sequence classification & sequence classification \\
  attention\_probs\_dropout\_prob & 0.1 & 0.1 & 0.1\\
  hidden\_act &gelu & gelu & gelu\\
  hidden\_dropout\_prob & 0.1 & 0.1 & 0.1\\
  hidden\_size & 1024 & 1024 & 768\\
  initializer\_range & 0.02 & 0.02 & 0.02\\
  intermediate\_size & 4096 & 4096 & 3072\\
  layer\_norm\_eps & 1e-12 & 1e-05 & 1e-12\\
  max\_position\_embeddings & 512 & 514 & 512\\
  model\_type & bert & roberta & bert\\
  num\_attention\_heads & 16 & 16 & 12\\
  num\_hidden\_layers & 24 & 24 & 12\\
  position\_embedding\_type & absolute & absolute & absolute\\
  vocab\_size & 30522 & 50265 & 31090\\
  learning\_rate & 2e-5& 2e-5& 2e-5\\
  epoch & 5 & 5 & 5\\
  \hline
    \end{tabular}
    \begin{tablenotes}
\item[a] \url{https://huggingface.co/bert-large-uncased}
\item[b] \url{https://huggingface.co/roberta-large}
\item[c] \url{https://github.com/allenai/scibert}
\end{tablenotes}
\end{threeparttable}
}
    \caption{Architecture and hyper parameters information. Our proposed hdBERT model ensembles RoBERTa and SciBERT via three MLP layers.}
    \label{tab:impts}
\end{table*}

\begin{table}
    \centering
    \begin{tabular}{l|c}
    \hline
        \bm{\mathrm{Statistical Information}} & \bm{\mathrm{SciAD}}$_{\mathrm{BI}}$ \\
         \hline
         number of training & 352,366\\
         number of development & 28,286\\
         number of test & 28,364\\
         \hline
    \end{tabular}
    \caption{The statistical information of SciAD$_{\mathrm{BI}}$ dataset.}
    \label{tab:newdataset}
\end{table}
As mentioned previously, we convert the original SciAD dataset into a binary classification dataset named SciAD$_{\mathrm{BI}}$ during modeling. 
For a sentence $\bm{s}$ with acronym $w_p$, $y=1$ if a long form $\bm{l}$ is true for $w_p$, while $y=0$ for other false candidate long forms of $w_p$. Specifically, to alleviate the imbalance problem during training, we upsample each positive sample to equal the number of candidate long forms of its acronym. More statistics of SciAD$_{\mathrm{BI}}$ is shown in Table~\ref{tab:newdataset}. We finally evaluate performances on SciAD's test dataset.

\subsection{Compared Methods}
We compare with several state-of-the-art and representative methods including Non-deep learning methods and Deep learning methods to verify the effectiveness of our proposed method.

\noindent \textbf{Non-deep learning methods}.
\begin{itemize}
    \item \textbf{MF}: most frequent which takes the long form with the highest frequency among all possible meanings of an acronym as the expanded form of the acronym.
    \item \textbf{ADE}~\cite{li2018guess}: a feature-based model that employs hand crafted features from the context of the acronyms to train a disambiguation classifier.
\end{itemize}
\noindent \textbf{Deep learning methods}.
\begin{itemize}
    \item \textbf{NOA}~\cite{charbonnier2018using} and  \textbf{UAD}~\cite{ciosici2019unsupervised}: language-model-based baselines that train the word embeddings using the training corpus.
    \item \textbf{DECBAE}~\cite{jin2019deep} and \textbf{BEM}~\cite{blevins2020moving}: models employing deep architectures (e.g., LSTM).
    \item \textbf{GAD}~\cite{veyseh-et-al-2020-what}: supervised method which utilizes syntactic structure of sentences to extend ambiguous acronyms in sentences by combining BiLSTM with GCN.
    \item \textbf{BERT}~\cite{devlin2018bert},  \textbf{RoBERTa}~\cite{liu2019roberta} and \textbf{SciBERT}~\cite{Beltagy2019SciBERT}: pretrained models use the now ubiquitous transformer architecture.
\end{itemize}

\subsection{Evaluation Metrics}
To evaluate the performance of different methods, three popular metrics are adopted, namely \textbf{Macro Precision}, \textbf{Macro Recall} and \textbf{Macro F1}. The definitions are as follows:
\begin{equation}
    \bm{\mathrm{Precision}}_\mathrm{MACRO} = \frac{\sum_{i=1}^{n}\mathrm{Precision}_i}{n}
\end{equation}
\begin{equation}
    \bm{\mathrm{Recall}}_\mathrm{MACRO} = \frac{\sum_{i=1}^{n}\mathrm{Recall}_i}{n}
\end{equation}
\begin{equation}
    \bm{\mathrm{F1}}_\mathrm{MACRO} = \frac{2\times\mathrm{Precision}_\mathrm{MACRO}\times\mathrm{Recall}_\mathrm{MACRO}}{\mathrm{Precision}_\mathrm{MACRO} + \mathrm{Recall}_\mathrm{MACRO}}
\end{equation}
where $n$ is the number of total classes, $\mathrm{Precision}_i$ and $\mathrm{Recall}_i$ represent the precision and recall of class $i$ respectively. The higher $\mathrm{Precision}_\mathrm{MACRO}$, $\mathrm{Recall}_\mathrm{MACRO}$ and $\mathrm{F1}_\mathrm{MACRO}$ indicate the higher performance of approaches.

\begin{table*}%[!t]
    \centering
    \begin{tabular}{l|ccc}
    \hline
         \textbf{Methodology} & \textbf{Macro Precision}(\%) & \textbf{Macro Recall}(\%) & \textbf{Macro F1}(\%) \\
         \hline
         \textbf{MF} & 89.03 & 42.20 & 57.26 \\
         \textbf{ADE}~\cite{li2018guess} & 86.74  & 43.25  & 57.72 \\
         \textbf{NOA}~\cite{charbonnier2018using} & 78.14  & 35.06  & 48.40 \\
         \textbf{UAD}~\cite{ciosici2019unsupervised} & 89.01  & 70.08  & 78.37 \\
         \textbf{BEM}~\cite{blevins2020moving} & 86.75  & 35.94  & 50.82 \\
         \textbf{DECBAE}~\cite{jin2019deep} & 88.67  & 74.32  & 80.86 \\
         \textbf{GAD}~\cite{veyseh-et-al-2020-what} & 89.27  & 76.66  & 81.90 \\
         \textbf{Human Performance}~\cite{veyseh-et-al-2020-what} & 97.82  & 94.45  & 96.10 \\
         \hline
         \textbf{MF} & 89.00 & 46.36 & 60.97 \\
         \textbf{BERT}~\cite{devlin2018bert} & 95.26& 86.92& 90.90\\
         \textbf{RoBERTa}~\cite{liu2019roberta} & 95.96 & 88.36 & 92.00\\
         \textbf{SciBERT}~\cite{Beltagy2019SciBERT} & 96.36 & 89.77 & 92.95\\
         \textbf{hdBERT}~(ours) & \textbf{96.94} & \textbf{90.73} & \textbf{93.73}\\
         \hline
    \end{tabular}
    \caption{Performance of models in acronym disambiguation.}
    \label{tab:results}
\end{table*}

\subsection{Implementation Details}
For models ADE, NOA, UAD, DECBAE, BEM, and GAD, please refer to~\citeauthor{veyseh-et-al-2020-what} for more implementation information.
We implement the proposed model based on Pytorch~\cite{paszke2019pytorch} and Transformers~\cite{wolf-etal-2020-transformers}.
For models BERT, RoBERTa, and SciBERT, we fine-tune them on dataset based on their popular pretrained models. The implementation details of these models are shown in Table~\ref{tab:impts}. Moreover, the information distillation components of our model are the same as model RoBERTa and SciBERT respectively. And we simply adopt three MLP layers for integration simultaneously. As mentioned previously, in the testing phase, the long form with the highest prediction probability in the candidate long form set of a sentence would be chosen as its final result. In addition, we use two V100 GPUs with 12 cores to complete all these experiments.

\subsection{Performance Comparison}

\begin{figure}
    \centering
    \includegraphics[width=\linewidth]{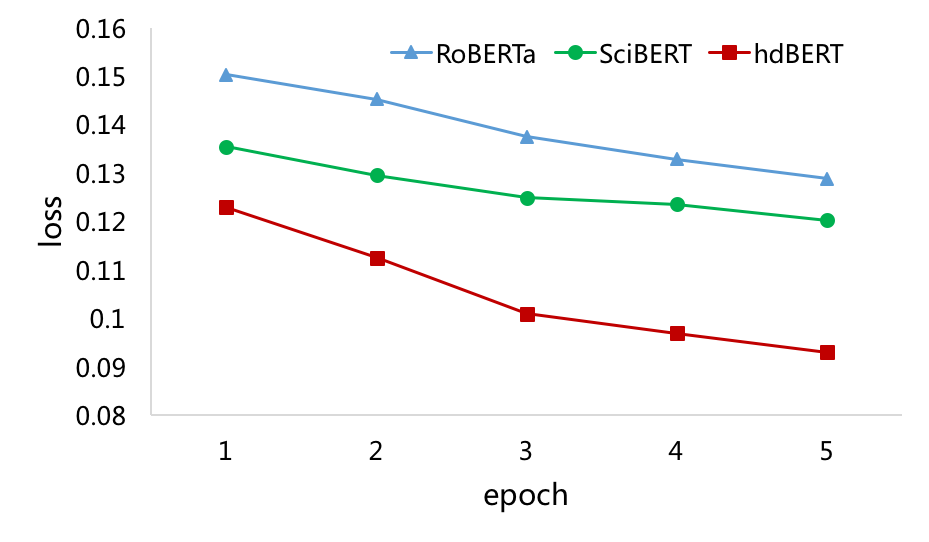}
    \caption{Loss curve on development dataset.}
    \label{fig:loss}
\end{figure}

Table~\ref{tab:results} demonstrates the main results of all compared methods~\footnote{We assume that both~\citeauthor{veyseh-et-al-2020-what} and this task have the same distribution of dataset due to the randomly dividing by the same ratio, making all these methods comparable.} on the dataset.
The major findings from the experimental results can be summarized as follows:

First, GAD achieves a better result than methods such as ADE, NOA, UAD, BEM, and DECBAE, showing the importance of syntactic structure for the acronym disambiguation task. But it still far worse than pretraining-based models like BERT and RoBERTa.
Second, between the two general-domain models, RoBERTa gets better performance than BERT, indicating the advantage of more fine-grained encoding. Moreover, SciBERT is more advanced than the domain agnostic methods, i.e., BERT and RoBERTa, with about 2.26\% and 1.03\% increased macro F1 respectively, showing the importance of the scientific domain pretraining for this task.
Furthermore, we can clearly observe that our hdBERT model outperforms all the baselines by a large margin. 
Its macro F1, with the reported value of 93.73\%, is about 1.88\% and 0.84\% higher than state-of-the-art RoBERTa and SciBERT respectively.
And its loss curve falls faster and converges lower than the two pretrained methods on the development dataset, as shown in Figure~\ref{fig:loss}.
These observations demonstrate that it is effective to model both fine-grained domain agnostic and high-level domain specific knowledge simultaneously.

However, despite the significant improvements among these approaches, performances of all models are still not as effective as humans on the dataset, especially on macro recall and macro F1, thus providing many further research opportunities for this scenario.

\subsection{Case Study}
\begin{table*}
    \centering
    	\resizebox{\linewidth}{!}{
    \begin{tabular}{l|l}
    \hline
        \textbf{Sentence} & \textbf{Conflicted Annotation} \\
        \hline
        Just like \textbf{RF}, QRF is a set of binary regression trees. & TR-43200: Regression Forest \\
        & TR-49535: Regression Function \\
        \hline
        Extensions of the \textbf{SBM} regarding the type of graph are reviewed in Section. & TR-17276: Sequential Monte Carlo\\
        &TR-47761: Stochastic Block Model\\
        \hline
        The obfuscated term is the term for which the \textbf{MACS} score is the lowest. & TR-15480: Mean Average Conceptual Similarity\\
        &TR-27970: Minimum Average Conceptual Similarity \\
        \hline
    \end{tabular}
    }
    \caption{Examples of noise data of SciAD dataset.}
    \label{tab:noise}
\end{table*}

We further focus on studying both success and failure cases of pretraining-based models to provide more insight into acronym disambiguation. 
Specifically, for success case of our model in which RoBERTa and SciBERT fail,
e.g., ``\textit{Each} \textit{SP} \textit{within} \textit{an} \textit{\textbf{SM}} \textit{shares} \textit{an} \textit{instruction} \textit{unit,} \textit{dedicated} \textit{to} \textit{the} \textit{management} \textit{of} \textit{the} \textit{instruction} \textit{flow} \textit{of} \textit{the} \textit{threads.}'' (DEV-6156), the true long form of ``\textit{SM}'' is
``\textit{Streaming} \textit{Multiprocessors}''. While both RoBERTa and SciBERT output ``\textit{Shared} \textit{Memory}'', which may often appear in deep learning publications. 
It might benefit from the additional integration modeling of two different information from RoBERTa and SciBERT.
However, all the three models fail in this example: ``\textit{In} \textit{the} \textit{first} \textit{stage,} \textit{we} \textit{train} \textit{the} \textit{SPM,} \textit{and} \textit{extract} \textit{the} \textit{\textbf{FL}} \textit{and} \textit{FR.}'' (DEV-4604) with the wrong prediction ``\textit{Federated} \textit{Learning}'' for ``\textit{FL}''. The true long form of ``\textit{FL}'' is ``\textit{Fixated} \textit{Locations}''. 
We guess that all models pay too much attention to ``\textit{Federated} \textit{Learning}'', a hot phrase nowadays, and ignore the subtle information among the sentence and its different candidate long forms.
It also indicates the necessity of more advanced models for this task.

\subsection{Further Discussion}
As mentioned previously and shown in Table~\ref{tab:results}, all the current models are still less effective than humans in this scenario. There are still many samples that all models fail in. Some further research opportunities on this dataset are discussed in this section. First, as shown in Table~\ref{tab:noise}, there are some noise data, i.e., conflicted annotation, in the SciAD dataset.
For example, the acronym ``\textit{RF}'' in boldface in sentence ``\textit{Just} \textit{like} \textit{\textbf{RF},} \textit{QRF} \textit{is} \textit{a} \textit{set} \textit{of} \textit{binary} \textit{regression} \textit{trees.}'' gets two different long form ``\textit{Regression} \textit{Forest}'' (TR-43200) and ``\textit{Regression} \textit{Function}'' (TR-49535) respectively. 
It will be some negative impacts on modeling to some extent.
Furthermore, to a certain extent, samples constructed from the same sentence with different long forms are independent during our training stage.
It might lose more subtle information among them.
Therefore, recent methods such as self-training~\cite{peng2019trainable,chi2020learning}, adversarial learning~\cite{fgsm,fgm,danqing2020}, and contrastive learning~\cite{hadsell2006dimensionality} are worth studying to further improve the performance.
\section{Conclusions}
\label{sec:conclusion}
An obstacle to scientific document understanding is the widespread use of acronyms which are shortened forms of long technical phrases.
Acronym disambiguation aims to find the correct meaning of an ambiguous acronym in a given text.
However, it is challenging and expensive to obtain large-scale high-quality annotated data in the scientific domain.
In this paper, we present a hierarchical dual-path BERT method coined hdBERT for acronym disambiguation to resolve the special challenges in this scenario.
The method is equipped with pretrained models RoBERTa and SciBERT and integrates their dual-path representations simultaneously to leveraging domain agnostic and specific knowledge.
Experiments on real-world datasets demonstrate the effectiveness of the proposed approach.
It achieves competitive performance and outperforms state-of-the-art methods among various evaluation metrics.
Moreover, there are still many research opportunities in this task, approaches such as self-training, adversarial learning, and contrastive learning are worth studying to further improve the performance.
\section{Acknowledgments}
\label{sec:acknowledgments}
We thank the organizers of acronym identification and disambiguation competitions and the reviewers for their valuable comments and suggestions.

% \clearpage
\bibliography{references}

\begin{thebibliography}{30}
\providecommand{\natexlab}[1]{#1}
\providecommand{\url}[1]{\texttt{#1}}
\providecommand{\urlprefix}{URL }
\expandafter\ifx\csname urlstyle\endcsname\relax
  \providecommand{\doi}[1]{doi:\discretionary{}{}{}#1}\else
  \providecommand{\doi}{doi:\discretionary{}{}{}\begingroup
  \urlstyle{rm}\Url}\fi

\bibitem[{Agirre, L{\'o}pez~de Lacalle, and Soroa(2014)}]{agirre2014random}
Agirre, E.; L{\'o}pez~de Lacalle, O.; and Soroa, A. 2014.
\newblock Random Walks for Knowledge-based Word Sense Disambiguation.
\newblock \emph{Computational Linguistics} 40(1): 57--84.

\bibitem[{Ammar et~al.(2018)Ammar, Groeneveld, Bhagavatula, Beltagy, Crawford,
  Downey, Dunkelberger, Elgohary, Feldman, Ha et~al.}]{ammar2018construction}
Ammar, W.; Groeneveld, D.; Bhagavatula, C.; Beltagy, I.; Crawford, M.; Downey,
  D.; Dunkelberger, J.; Elgohary, A.; Feldman, S.; Ha, V.; et~al. 2018.
\newblock Construction of the Literature Graph in Semantic Scholar.
\newblock In \emph{NAACL}, 84--91.

\bibitem[{Barba et~al.(2020)Barba, Procopio, Campolungo, Pasini, and
  Navigli}]{barba2020mulan}
Barba, E.; Procopio, L.; Campolungo, N.; Pasini, T.; and Navigli, R. 2020.
\newblock MuLaN: Multilingual Label Propagation for Word Sense Disambiguation.
\newblock In \emph{IJCAI}, 3837--3844.

\bibitem[{Beltagy, Lo, and Cohan(2019)}]{Beltagy2019SciBERT}
Beltagy, I.; Lo, K.; and Cohan, A. 2019.
\newblock SciBERT: A Pretrained Language Model for Scientific Text.
\newblock In \emph{EMNLP}, 3606--3611.

\bibitem[{Bevilacqua and Navigli(2020)}]{bevilacqua2020breaking}
Bevilacqua, M.; and Navigli, R. 2020.
\newblock Breaking Through the 80\% Glass Ceiling: Raising the State of the Art
  in Word Sense Disambiguation by Incorporating Knowledge Graph Information.
\newblock In \emph{ACL}, 2854--2864.

\bibitem[{Blevins and Zettlemoyer(2020)}]{blevins2020moving}
Blevins, T.; and Zettlemoyer, L. 2020.
\newblock Moving Down the Long Tail of Word Sense Disambiguation with
  Gloss-Informed Biencoders.
\newblock \emph{arXiv preprint arXiv:2005.02590} .

\bibitem[{Charbonnier and Wartena(2018)}]{charbonnier2018using}
Charbonnier, J.; and Wartena, C. 2018.
\newblock Using Word Embeddings for Unsupervised Acronym Disambiguation.
\newblock In \emph{COLING}, 2610--2619.

\bibitem[{Chi et~al.(2020)Chi, Zeng, Zhong, Liang, Feng, Ao, and
  Tang}]{chi2020learning}
Chi, J.; Zeng, G.; Zhong, Q.; Liang, T.; Feng, J.; Ao, X.; and Tang, J. 2020.
\newblock Learning to Undersampling for Class Imbalanced Credit Risk
  Forecasting.
\newblock In \emph{ICDM}.

\bibitem[{Ciosici, Sommer, and Assent(2019)}]{ciosici2019unsupervised}
Ciosici, M.; Sommer, T.; and Assent, I. 2019.
\newblock Unsupervised Abbreviation Disambiguation Contextual Disambiguation
  using Word Embeddings.
\newblock \emph{arXiv preprint arXiv:1904.00929} .

\bibitem[{Devlin et~al.(2019)Devlin, Chang, Lee, and
  Toutanova}]{devlin2018bert}
Devlin, J.; Chang, M.; Lee, K.; and Toutanova, K. 2019.
\newblock BERT: Pre-training of Deep Bidirectional Transformers for Language
  Understanding.
\newblock In \emph{NAACL}, 4171--4186.

\bibitem[{Goodfellow, Shlens, and Szegedy(2015)}]{fgsm}
Goodfellow, I.~J.; Shlens, J.; and Szegedy, C. 2015.
\newblock Explaining and Harnessing Adversarial Examples.
\newblock In \emph{ICLR}.

\bibitem[{Hadsell, Chopra, and LeCun(2006)}]{hadsell2006dimensionality}
Hadsell, R.; Chopra, S.; and LeCun, Y. 2006.
\newblock Dimensionality Reduction by Learning an Invariant Mapping.
\newblock In \emph{CVPR}, volume~2, 1735--1742.

\bibitem[{Jin, Liu, and Lu(2019)}]{jin2019deep}
Jin, Q.; Liu, J.; and Lu, X. 2019.
\newblock Deep Contextualized Biomedical Abbreviation Expansion.
\newblock In \emph{BioNLP Workshop}, 88--96.

\bibitem[{Lee et~al.(2020)Lee, Yoon, Kim, Kim, Kim, So, and
  Kang}]{lee2020biobert}
Lee, J.; Yoon, W.; Kim, S.; Kim, D.; Kim, S.; So, C.~H.; and Kang, J. 2020.
\newblock BioBERT: A Pre-trained Biomedical Language Representation Model for
  Biomedical Text Mining.
\newblock \emph{Bioinformatics} 36(4): 1234--1240.

\bibitem[{Li et~al.(2018)Li, Zhao, Fuxman, and Tao}]{li2018guess}
Li, Y.; Zhao, B.; Fuxman, A.; and Tao, F. 2018.
\newblock Guess Me if You Can: Acronym Disambiguation for Enterprises.
\newblock In \emph{ACL}, 1308--1317.

\bibitem[{Liu et~al.(2019)Liu, Ott, Goyal, Du, Joshi, Chen, Levy, Lewis,
  Zettlemoyer, and Stoyanov}]{liu2019roberta}
Liu, Y.; Ott, M.; Goyal, N.; Du, J.; Joshi, M.; Chen, D.; Levy, O.; Lewis, M.;
  Zettlemoyer, L.; and Stoyanov, V. 2019.
\newblock Roberta: A Robustly Optimized BERT Pretraining Approach.
\newblock \emph{arXiv preprint arXiv:1907.11692} .

\bibitem[{Miyato, Dai, and Goodfellow(2017)}]{fgm}
Miyato, T.; Dai, A.~M.; and Goodfellow, I. 2017.
\newblock Adversarial Training Methods for Semi-supervised Text Classification.
\newblock In \emph{ICLR}.

\bibitem[{Navigli(2009)}]{navigli2009word}
Navigli, R. 2009.
\newblock Word Sense Disambiguation: A Survey.
\newblock \emph{ACM computing surveys (CSUR)} 41(2): 1--69.

\bibitem[{Paszke et~al.(2019)Paszke, Gross, Massa, Lerer, Bradbury, Chanan,
  Killeen, Lin, Gimelshein, Antiga et~al.}]{paszke2019pytorch}
Paszke, A.; Gross, S.; Massa, F.; Lerer, A.; Bradbury, J.; Chanan, G.; Killeen,
  T.; Lin, Z.; Gimelshein, N.; Antiga, L.; et~al. 2019.
\newblock Pytorch: An Imperative Style, High-performance Deep Learning Library.
\newblock In \emph{NIPS}, 8026--8037.

\bibitem[{Peng et~al.(2019)Peng, Zhang, Xing, Gui, Huang, Jiang, Ding, and
  Chen}]{peng2019trainable}
Peng, M.; Zhang, Q.; Xing, X.; Gui, T.; Huang, X.; Jiang, Y.-G.; Ding, K.; and
  Chen, Z. 2019.
\newblock Trainable Undersampling for Class-imbalance Learning.
\newblock In \emph{AAAI}, volume~33, 4707--4714.

\bibitem[{Prokofyev et~al.(2013)Prokofyev, Demartini, Boyarsky, Ruchayskiy, and
  Cudr{\'e}-Mauroux}]{prokofyev2013ontology}
Prokofyev, R.; Demartini, G.; Boyarsky, A.; Ruchayskiy, O.; and
  Cudr{\'e}-Mauroux, P. 2013.
\newblock Ontology-based Word Sense Disambiguation for Scientific Literature.
\newblock In \emph{ECIR}, 594--605.

\bibitem[{Scarlini, Pasini, and Navigli(2020)}]{scarlini2020sensembert}
Scarlini, B.; Pasini, T.; and Navigli, R. 2020.
\newblock SensEmBERT: Context-Enhanced Sense Embeddings for Multilingual Word
  Sense Disambiguation.
\newblock In \emph{AAAI}, 8758--8765.

\bibitem[{Sennrich, Haddow, and Birch(2016)}]{sennrich2016neural}
Sennrich, R.; Haddow, B.; and Birch, A. 2016.
\newblock Neural Machine Translation of Rare Words with Subword Units.
\newblock In \emph{ACL}, 1715--1725.

\bibitem[{Vaswani et~al.(2017)Vaswani, Shazeer, Parmar, Uszkoreit, Jones,
  Gomez, Kaiser, and Polosukhin}]{vaswani2017attention}
Vaswani, A.; Shazeer, N.; Parmar, N.; Uszkoreit, J.; Jones, L.; Gomez, A.~N.;
  Kaiser, {\L}.; and Polosukhin, I. 2017.
\newblock Attention is All you Need.
\newblock In \emph{NIPS}, 5998--6008.

\bibitem[{Veyseh et~al.(2020{\natexlab{a}})Veyseh, Dernoncourt, Nguyen, Chang,
  and Celi}]{veyseh2020acronym}
Veyseh, A. P.~B.; Dernoncourt, F.; Nguyen, T.~H.; Chang, W.; and Celi, L.~A.
  2020{\natexlab{a}}.
\newblock Acronym Identification and Disambiguation shared tasks for Scientific
  Document Understanding.
\newblock In \emph{AAAI Workshop on Scientific Document Understanding}.

\bibitem[{Veyseh et~al.(2020{\natexlab{b}})Veyseh, Dernoncourt, Tran, and
  Nguyen}]{veyseh-et-al-2020-what}
Veyseh, A. P.~B.; Dernoncourt, F.; Tran, Q.~H.; and Nguyen, T.~H.
  2020{\natexlab{b}}.
\newblock What Does This Acronym Mean? Introducing a New Dataset for Acronym
  Identification and Disambiguation.
\newblock In \emph{COLING}, 3285--3301.

\bibitem[{Wang, Wang, and Fujita(2020)}]{wang2020word}
Wang, Y.; Wang, M.; and Fujita, H. 2020.
\newblock Word Sense Disambiguation: A Comprehensive Knowledge Exploitation
  Framework.
\newblock \emph{Knowledge-Based Systems} 190: 105030.

\bibitem[{Wolf et~al.(2020)Wolf, Chaumond, Debut, Sanh, Delangue, Moi, Cistac,
  Funtowicz, Davison, Shleifer et~al.}]{wolf-etal-2020-transformers}
Wolf, T.; Chaumond, J.; Debut, L.; Sanh, V.; Delangue, C.; Moi, A.; Cistac, P.;
  Funtowicz, M.; Davison, J.; Shleifer, S.; et~al. 2020.
\newblock Transformers: State-of-the-art Natural Language Processing.
\newblock In \emph{EMNLP}, 38--45.

\bibitem[{Wu et~al.(2016)Wu, Schuster, Chen, Le, Norouzi, Macherey, Krikun,
  Cao, Gao, Macherey et~al.}]{wu2016google}
Wu, Y.; Schuster, M.; Chen, Z.; Le, Q.~V.; Norouzi, M.; Macherey, W.; Krikun,
  M.; Cao, Y.; Gao, Q.; Macherey, K.; et~al. 2016.
\newblock Google's Neural Machine Translation System: Bridging the Gap between
  Human and Machine Translation.
\newblock \emph{arXiv preprint arXiv:1609.08144} .

\bibitem[{Zhu et~al.(2021)Zhu, Lin, Zhang, Zhong, Zeng, Wu, and
  Tang}]{danqing2020}
Zhu, D.; Lin, W.; Zhang, Y.; Zhong, Q.; Zeng, G.; Wu, W.; and Tang, J. 2021.
\newblock AT-BERT: Adversarial Training BERT for Acronym Identification.
\newblock In \emph{AAAI Workshop on Scientific Document Understanding}.

\end{thebibliography}

\end{document}